\documentclass[conference]{IEEEtran}

\usepackage[utf8]{inputenc}
\usepackage[T1]{fontenc}
\usepackage{amsmath,amssymb}
\usepackage{graphicx}
\usepackage{hyperref}
\usepackage{cite}
\usepackage{booktabs}
\usepackage{algorithm}
\usepackage{algorithmic}
\usepackage{tikz}
\usetikzlibrary{positioning, arrows.meta, calc, fit, shapes.geometric, decorations.pathreplacing}
\usepackage{pgfplots}
\pgfplotsset{compat=1.18}
\usepgfplotslibrary{groupplots}
\usepackage[caption=false]{subfig}

\definecolor{myred}{HTML}{9E292B}
\definecolor{myblue}{HTML}{235787}
\definecolor{mygreen}{HTML}{5E6638}
\definecolor{mygray}{HTML}{444444}
\definecolor{myblack}{HTML}{000000}
\definecolor{mywhite}{HTML}{FFFFFF}

\definecolor{myaltred}{HTML}{D46A78}
\definecolor{myaltblue}{HTML}{6699C2}
\definecolor{myaltgreen}{HTML}{B0B58C}
\definecolor{myaltgray}{HTML}{AAAAAA}

\definecolor{mydarkred}{HTML}{741E1A}
\definecolor{mylightred1}{HTML}{B15455}
\definecolor{mylightred2}{HTML}{C57F80}
\definecolor{mylightred3}{HTML}{D8A9AA}
\definecolor{mylightred4}{HTML}{ECD4D5}
\definecolor{mylightblue1}{HTML}{5A7DA5}
\definecolor{mylightblue2}{HTML}{7D99BA}
\definecolor{mylightblue3}{HTML}{B3C3D7}
\definecolor{mylightblue4}{HTML}{D3DCE8}
\definecolor{mydarkgreen}{HTML}{3E4822}
\definecolor{mylightgreen1}{HTML}{828859}
\definecolor{mylightgreen2}{HTML}{9AA075}
\definecolor{mylightgreen3}{HTML}{B8BC96}
\definecolor{mylightgreen4}{HTML}{D4D4B8}
\definecolor{mylightgray1}{HTML}{6F6F6F}
\definecolor{mylightgray2}{HTML}{999999}
\definecolor{mylightgray3}{HTML}{B4B4B4}
\definecolor{mylightgray4}{HTML}{DCDCDC}

\definecolor{nvidiagreen}{HTML}{76B900}
\definecolor{nvidiapurple}{HTML}{952FC6}
\definecolor{nvidiaorange}{HTML}{EF9100}
\definecolor{nvidiamagenta}{HTML}{D2308E}
\definecolor{nvidiateal}{HTML}{1DBBA4}
\definecolor{nvidiagray}{HTML}{757575}

\definecolor{greenlight2}{HTML}{CFFF40}
\definecolor{greenlight1}{HTML}{BFF230}
\definecolor{greendark1}{HTML}{3F8500}
\definecolor{greendark2}{HTML}{265600}

\definecolor{purplelight2}{HTML}{F9D4FF}
\definecolor{purplelight1}{HTML}{C359EF}
\definecolor{purpledark1}{HTML}{741D9D}
\definecolor{purpledark2}{HTML}{4D1368}

\definecolor{orangelight2}{HTML}{FEEEB2}
\definecolor{orangelight1}{HTML}{FCDE7B}
\definecolor{orangedark1}{HTML}{EF9100}
\definecolor{orangedark2}{HTML}{DF6500}

\definecolor{magentalight2}{HTML}{FFD3F2}
\definecolor{magentalight1}{HTML}{FC79CA}
\definecolor{magentadark1}{HTML}{8C1C55}
\definecolor{magentadark2}{HTML}{5D1337}

\definecolor{teallight2}{HTML}{ADFCF8}
\definecolor{teallight1}{HTML}{9AEFE5}
\definecolor{tealdark1}{HTML}{0D8473}
\definecolor{tealdark2}{HTML}{04554B}

\tikzset{
  labelstyle/.style={
    fill=white, fill opacity=0.85, text opacity=1, inner sep=0.2mm, rotate=0, font=\sffamily\footnotesize
  },
}

\pgfplotsset{
ber_plotstyle/.style={
width=6cm, 
height=3.5cm,
scale only axis,
axis background/.style={fill=none},
axis x line*=bottom,  
axis y line*=left,  
clip mode=individual,
every outer y axis line/.append style={black, very thick, cap=rect},
every outer x axis line/.append style={black, very thick, cap=rect},
ymax=1,
xmajorgrids,
ymajorgrids,
yminorgrids,
xlabel={SNR [dB]},
ylabel style = {black, font=\small, yshift=0.1cm},
xlabel style = {black, font=\small, yshift=-0.1cm},
yticklabel style = {font=\footnotesize, text width=0.5cm, align=left},
xticklabel style = {font=\footnotesize, black},
tick align=outside,
tick style={black, very thick},
line width={2pt},
major tick length=2.25pt, 
minor tick length=0pt,
xtick pos=bottom,      
ytick pos=left,
}
}

\pgfplotsset{
standard_plotstyle/.style={
width=6cm, 
height=3.5cm,
scale only axis,
axis background/.style={fill=none},
axis x line*=bottom,  
axis y line*=left,  
clip mode=individual,
every outer y axis line/.append style={black, very thick, cap=rect},
every outer x axis line/.append style={black, very thick, cap=rect},
xmajorgrids,
ymajorgrids,
yminorgrids,
ylabel style = {black, font=\small, yshift=0.cm},
xlabel style = {black, font=\small, yshift=0.cm},
yticklabel style = {font=\footnotesize\rmfamily, text width=0.5cm, align=right},
xticklabel style = {font=\footnotesize, black},
tick align=outside,
tick style={black, very thick},
line width={2pt},
major tick length=2.25pt, 
minor tick length=0pt,
xtick pos=bottom,      
ytick pos=left,
}
}

\title{The AI Telco Engineer: Toward Autonomous Discovery of Wireless Communications Algorithms}

\author{
\IEEEauthorblockN{Fayçal Aït Aoudia, Jakob Hoydis, Sebastian Cammerer,\\
Lorenzo Maggi, Gian Marti, and Alexander Keller}
}

\begin{document}

\maketitle
\renewcommand{\thefootnote}{\fnsymbol{footnote}}
\footnotetext[1]{All authors are with NVIDIA. Contact: \texttt{faitaoudia@nvidia.com}. Repository: \url{https://github.com/nvlabs/the-ai-telco-engineer}.}
\renewcommand{\thefootnote}{\arabic{footnote}}

\begin{abstract}
Agentic AI is rapidly transforming the way research is conducted, from prototyping ideas to reproducing results found in the literature. In this paper, we explore the ability of agentic AI to autonomously design wireless communication algorithms. To that end, we implement a dedicated framework that leverages large language models (LLMs) to iteratively generate, evaluate, and refine candidate algorithms. We evaluate the framework on three tasks spanning the physical (PHY) and medium access control (MAC) layers: statistics-agnostic channel estimation, channel estimation with known covariance, and link adaptation. Our results show that, in a matter of hours, the framework produces algorithms that are competitive with and, in some cases, outperforming conventional baselines. Moreover, unlike neural network-based approaches, the generated algorithms are fully explainable and extensible. This work represents a first step toward the autonomous discovery of novel wireless communication algorithms, and we look forward to the progress our community makes in this direction.
\end{abstract}
\section{Introduction}
\label{sec:introduction}

The pace at which agentic AI capabilities are advancing has begun to fundamentally alter research workflows. With new agentic frameworks being released on a daily basis, many researchers have already integrated such tools into their workflow. In particular, coding agents such as Claude Code, GPT Codex, and Cursor make it easier than ever to rapidly prototype ideas, removing the burden of complex software toolchains that was traditionally associated with the development and evaluation of new algorithmic concepts.

Reproducing results from the literature has also become significantly more accessible thanks to dedicated agentic frameworks, such as DeepCode~\cite{li2025deepcodeopenagenticcoding}, which aim to precisely reproduce the results of a publication solely from the document itself. With such tools, there is little reason not to implement baselines when benchmarking new methods. One should also highlight deep search tools, which autonomously build research reports by exploring the literature based on a user query. Examples of such tools include GPT Researcher~\cite{Elovic_gpt-researcher_2023}, as well as the deep search capabilities offered by OpenAI, Anthropic Claude, Google Gemini, Perplexity, and others.

In this context, it is only natural to consider how agentic AI could impact research in wireless communications. We tackle this vast question by exploring the ability of agentic AI to autonomously design wireless communication algorithms. To that end, we have built a dedicated agentic AI framework, \emph{The AI Telco Engineer}, that leverages Sionna~\cite{sionna} to evaluate algorithms in link- and system-level simulations. The framework includes a tool for the agent to access the Sionna API documentation, developer guides, and tutorials. Code is executed in containerized environments, referred to as \emph{workspaces}, in which filesystem editing tools are also available to the agent. Using this framework, we show that agentic AI can design wireless communication algorithms that are competitive with, and in some cases outperform, conventional approaches.

Similar frameworks for program evolution exist, such as ShinkaEvolve~\cite{lange2025shinka} and OpenEvolve~\cite{openevolve}. However, we have opted to build a custom framework in order to freely experiment with agentic AI strategies best suited to wireless communication algorithm design. We plan to continue improving our framework to further specialize it for this domain and enhance its performance.

The remainder of this paper is organized as follows. Section~\ref{sec:framework_architecture} presents the architecture of the proposed framework. Section~\ref{sec:tasks_and_results} describes the considered tasks and discusses the obtained results. Section~\ref{sec:conclusion} concludes the paper.

\section{Framework Architecture}
\label{sec:framework_architecture}

\begin{figure}[t]
    \centering
    \begin{tikzpicture}[
    box/.style={thick, draw, minimum width=2.4cm, minimum height=1.4cm, align=center, font=\footnotesize\bfseries},
    lbl/.style={font=\footnotesize, align=center},
    arr/.style={thick, ->, >=stealth}
]
    \node[box] (fw) {The AI Telco\\Engineer};

    \node[lbl, left=1.2cm of fw, yshift=0.55cm]  (task) {Task\\description};
    \node[lbl, left=1.2cm of fw, yshift=-0.55cm] (eval) {Evaluation\\tool};

    \node[lbl, right=1.2cm of fw] (out) {Ranked set of\\algorithms};

    \draw[arr] (task.east) -- (task.east -| fw.west);
    \draw[arr] (eval.east) -- (eval.east -| fw.west);
    \draw[arr] (fw.east) -- (out.west);
\end{tikzpicture}
    \caption{The framework from an input/output perspective. The user provides a task, i.e., a problem description paired with an evaluation tool, and the framework returns a ranked set of algorithms.}
    \label{fig:io_view}
\end{figure}
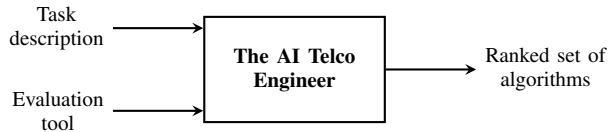

An input/output view of the framework is shown in Fig.~\ref{fig:io_view}. The framework takes as input a \emph{task}, which consists of a description of the problem for which algorithms are to be generated as well as an \emph{evaluation tool} used by the agent to assess a generated algorithm against a user-defined metric. To prevent metric gaming -- where the agent modifies the evaluation procedure to artificially inflate its score -- the evaluation tool is immutable; agents can only invoke it. As output, the framework produces not a single algorithm, but a set of algorithms ranked according to the user-defined metric.

Fundamentally, the framework builds on the idea of using LLMs to explore the space of algorithms solving a well-defined problem~\cite{Hemberg2024, liu2024llm4ad, novikov2025alphaevolve}, by iteratively expanding a population of candidate solutions. The remainder of this section details the architecture of the optimization loop, the agent workflow, and the tools available to agents.

\subsection{Overview}

\begin{figure}[t]
    \centering
    \begin{tikzpicture}[
    box/.style={draw, thick, minimum height=0.8cm, align=center, font=\footnotesize},
    orch/.style={box, thick, minimum width=2.2cm, font=\footnotesize\bfseries},
    idea/.style={minimum height=0.8cm, minimum width=1.0cm, align=center, font=\footnotesize},
    agent/.style={draw, thick, minimum width=0.9cm, minimum height=0.9cm, align=center, font=\scriptsize},
    arr/.style={->, thick, >=stealth},
    lbl/.style={font=\scriptsize}
]
    \node[orch] (orch) {Orchestrator};

    \node[idea, below left=1.2cm and 1.2cm of orch]  (i1) {Idea 1};
    \node[idea, below=1.2cm of orch]                  (i2) {Idea 2};
    \node[idea, below right=1.2cm and 1.2cm of orch]  (in) {Idea $N$};

    \node at ($(i2)!0.5!(in)$) {\footnotesize$\cdots$};

    \node[agent, below=0.8cm of i1, xshift=-0.55cm] (a1) {Agent\\1};
    \node[agent, below=0.8cm of i1, xshift= 0.55cm] (a2) {Agent\\2};

    \node[agent, below=0.8cm of i2, xshift=-0.55cm] (a3) {Agent\\3};
    \node[agent, below=0.8cm of i2, xshift= 0.55cm] (a4) {Agent\\4};
    
    \node[agent, below=0.8cm of in, xshift=-0.55cm] (an1) {Agent\\$M\!-\!1$};
    \node[agent, below=0.8cm of in, xshift= 0.55cm] (an2) {Agent\\$M$};

    \node at ($(a4)!0.5!(an1)$) {\footnotesize$\cdots$};

    \draw[arr] (orch.south) -- ([xshift=0.3cm, yshift=-0.05cm]i1.north);
    \draw[arr] (orch.south) -- (i2.north);
    \draw[arr] (orch.south) -- ([xshift=-0.3cm, yshift=-0.05cm]in.north);

    \draw[arr] (i1.south) -- (a1.north);
    \draw[arr] (i1.south) -- (a2.north);
    \draw[arr] (i2.south) -- (a3.north);
    \draw[arr] (i2.south) -- (a4.north);
    \draw[arr] (in.south) -- (an1.north);
    \draw[arr] (in.south) -- (an2.north);

    \draw[thick, decorate, decoration={brace, mirror, amplitude=5pt}]
        ([yshift=-0.2cm]a1.south west) -- ([yshift=-0.2cm]an2.south east)
        coordinate[midway, below=10pt] (brace-mid);

    \draw[arr, rounded corners=4pt]
        (brace-mid) -- ++(0, -0.3)
        -| ([xshift=-0.5cm]a1.west |- orch.west)
        -- (orch.west)
        node[lbl, above, pos=0.45] {Generated algorithms};
\end{tikzpicture}
    \caption{Overview of the idea-driven iterative optimization loop. The orchestrator generates ideas that are distributed to parallel agents, each operating in its own workspace. Algorithms flow back to the orchestrator to inform the next iteration. In this figure, each idea is assigned to two agents.}
    \label{fig:overview}
\end{figure}
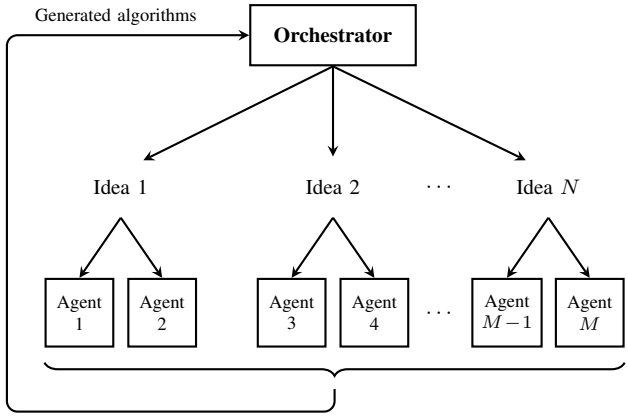

The framework follows a two-tier architecture: an \emph{orchestrator} LLM drives the optimization process, while a population of \emph{agent} LLMs implements and refines solutions in parallel. An overview of the architecture is shown in Fig.~\ref{fig:overview}.

The optimization proceeds over multiple iterations. At each iteration, the orchestrator proposes $N$ distinct algorithmic ideas, and a population of $M$ agents is distributed across these ideas, with $M/N$ agents assigned to each. Agents work independently and in parallel, each within its own isolated containerized environment referred to as a \emph{workspace}. Each idea is typically assigned to multiple agents to exploit LLM stochasticity: different agents assigned to the same idea will likely produce different implementation variants (as long as the LLM's temperature is set sufficiently high), effectively exploring multiple realizations of a single approach. This diversity reduces the risk of a promising idea being discarded due to a single poor implementation.

After an iteration completes, the orchestrator reviews the outcomes and proposes a new set of ideas for the next iteration. The orchestrator considers solutions from all previous iterations, not only the most recent one. To avoid saturating the context window, solutions produced by agents are first summarized in natural language to highlight their key aspects, and these summaries are provided to the orchestrator LLM alongside the corresponding metrics. Ideas proposed for subsequent iterations can be entirely novel approaches, refinements of previously explored ideas, or combinations of multiple prior ideas; the orchestrator decides how to proceed based on the accumulated results. The full optimization loop is summarized in Algorithm~\ref{alg:optimization}.

\begin{algorithm}[t]
\caption{Idea-driven iterative optimization}
\label{alg:optimization}
\begin{algorithmic}[1]
\REQUIRE Task description $\mathcal{T}$, evaluation tool $\mathcal{E}$, number of ideas $N$, population size $M$, number of iterations $G$
\STATE $\mathcal{S} \leftarrow \emptyset$
\FOR{$g = 1$ \TO $G$}
    \STATE $\triangleright$ \textit{$\mathcal{I}_g$: set of $N$ ideas for iteration $g$}
    \IF{$g = 1$}
        \STATE $\mathcal{I}_g \leftarrow \mathrm{Orchestrator}(\mathcal{T},\, N)$
    \ELSE
        \STATE $\mathcal{I}_g \leftarrow \mathrm{Orchestrator}(\mathcal{T},\, N,\, \mathrm{summaries}(\mathcal{S}))$
    \ENDIF
    \FORALL{idea $I_k \in \mathcal{I}_g$}
        \FOR{$j = 1$ \TO $M/N$}
            \STATE Spawn agent with $(\mathcal{T},\, I_k,\, \mathcal{E})$
        \ENDFOR
    \ENDFOR
    \STATE $\triangleright$ \textit{Run all agents in parallel}
    \FORALL{completed agent with solution $s$}
        \STATE $\mathrm{metric} \leftarrow \mathcal{E}(s)$
        \STATE $\mathrm{summary} \leftarrow \mathrm{Orchestrator.summarize}(s,\, I_k)$
        \STATE $\mathcal{S} \leftarrow \mathcal{S} \cup \{(s,\, \mathrm{metric},\, \mathrm{summary})\}$
    \ENDFOR
\ENDFOR
\RETURN $\mathcal{S}$ ranked by metric
\end{algorithmic}
\end{algorithm}

\subsection{Agent Workflow}

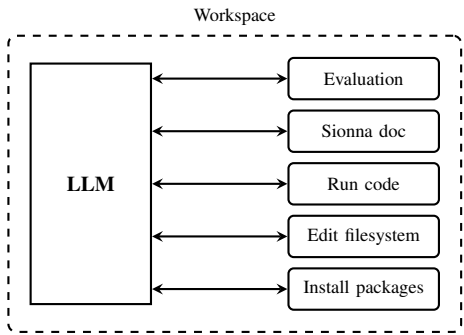
\begin{figure}[t]
    \centering
    \begin{tikzpicture}[
    agent/.style={thick, draw, minimum width=1.6cm, minimum height=3.2cm, align=center, font=\footnotesize\bfseries},
    tool/.style={thick, draw, rounded corners=2pt, minimum width=2cm, minimum height=0.55cm, align=center, font=\scriptsize},
    arr/.style={thick, <->, >=stealth},
]
    \node[agent] (agent) {LLM};

    \node[tool, right=1.8cm of agent, yshift=1.4cm]  (t1) {Evaluation};
    \node[tool, right=1.8cm of agent, yshift=0.7cm]  (t2) {Sionna doc};
    \node[tool, right=1.8cm of agent]                 (t3) {Run code};
    \node[tool, right=1.8cm of agent, yshift=-0.7cm] (t4) {Edit filesystem};
    \node[tool, right=1.8cm of agent, yshift=-1.4cm] (t5) {Install packages};

    \draw[arr] (agent.east |- t1) -- (t1.west);
    \draw[arr] (agent.east |- t2) -- (t2.west);
    \draw[arr] (agent.east |- t3) -- (t3.west);
    \draw[arr] (agent.east |- t4) -- (t4.west);
    \draw[arr] (agent.east |- t5) -- (t5.west);

    \node[draw, thick, dashed, rounded corners=3pt, inner sep=8pt,
          fit=(agent)(t1)(t5),
          label={[font=\scriptsize]above:Workspace}] {};
\end{tikzpicture}
    \caption{Agent runtime environment. Each agent is an LLM running inside an isolated containerized workspace with access to a set of tools.}
    \label{fig:agent_workflow}
\end{figure}

Each agent operates within an isolated Docker container and is given the task description together with an assigned algorithmic approach and, optionally, reference code from prior iterations. As illustrated in Fig.~\ref{fig:agent_workflow}, each agent is an LLM with access to a set of tools, all running inside a containerized workspace.

The agent follows a structured two-file workflow. A file \texttt{draft.py} serves as a scratch pad for experimentation: the agent writes, edits, and evaluates code in this file. A second file, \texttt{solution.py}, acts as a vault for the best result found so far. The agent is instructed to copy \texttt{draft.py} to \texttt{solution.py} only when a new best metric is achieved, ensuring that progress is preserved even if the agent is interrupted. The agent iteratively implements its approach, invokes the evaluation tool to measure performance, and refines the solution until it runs out of ideas or time. A configurable timeout bounds each agent's execution. At the end of each run, whether by normal completion or timeout, a post-run evaluation is automatically performed on the final \texttt{solution.py}, and the result is recorded.

Within its workspace, an agent can execute arbitrary code, carry out file management operations (read, write, edit, copy, and delete files), and install packages. A workspace only lives for the duration of a single iteration. At every iteration, a fresh workspace is spawned for each agent, ensuring that corruption of a workspace does not impact subsequent iterations.

\subsection{Sionna Documentation Tool}

The framework uses Sionna~\cite{sionna} to perform link-level and system-level simulations. To enable agents to write correct Sionna code, we provide them with tools to access the Sionna API documentation, tutorials, and developer guides through three operations. \emph{Search} performs semantic search over the documentation using a FAISS~\cite{douze2025faiss} vector store with a cross-encoder reranker; the vector store is built once from Sionna API docstrings and online tutorials, then memory-mapped across worker processes. \emph{Help} retrieves the full docstring and signature of any Sionna symbol by executing introspection code within the workspace. \emph{List} enumerates available classes and functions within a given Sionna module. These tools allow agents to use Sionna blocks and utilities, and to understand Sionna data structures such as \texttt{ResourceGrid} and \texttt{StreamManagement} when required.

\subsection{Evaluation Tool}

A task-specific evaluation tool is provided as a pluggable component by the user. It executes the agent's code and returns a structured output with three components. First, a binary success/failure flag indicates whether the evaluation completed successfully. Because agents generate code autonomously, execution failures such as runtime errors or incompatible outputs are common, particularly in early iterations. A failure can also be triggered by the evaluation tool itself when a constraint is not satisfied, as discussed below. In case of failure, the error stack trace or a diagnostic message is returned to help the agent debug or refine its code. Second, a scalar primary metric serves as the optimization objective. Third, the evaluation tool may return optional auxiliary information, such as additional metrics or diagnostic data, which is provided to the agent as feedback but is not used for ranking. This design accommodates problems with multiple objectives: the evaluation tool can encode constraints into the success flag, for instance, marking a run as failed if a secondary metric violates a prescribed threshold, so that the primary metric is optimized only over the feasible set. As noted above, the evaluation tool is not editable by the agents, preventing metric gaming. Moreover, each call to the evaluation tool is subject to a timeout, which prevents any single evaluation from blocking the optimization loop and also serves as an implicit constraint on the computational complexity of the generated algorithms.

\section{Experimental Evaluation}
\label{sec:tasks_and_results}

To evaluate the framework, we consider three tasks: two PHY layer tasks involving channel estimation, and one MAC layer task involving link adaptation. We emphasize that the framework is not specific to these tasks, as it could be applied to any wireless communication problem for which an evaluation tool can be defined.

For all tasks, two off-the-shelf LLMs are considered: a closed-weights model, GPT~5.4, and an open-weights model, GPT-OSS~120B. The framework was run once per model for each task, resulting in two independent sets of generated algorithms per task. Both the orchestrator and the agents were configured to use the same model within a given run. The framework was configured to execute $G = 10$ iterations, each with $M = 20$ agents running in parallel and the orchestrator generating $N = 5$ ideas per iteration. Each idea was therefore assigned to $4$ agents. The timeout per agent was set to $20$~minutes, and agents had access to GPUs for executing Python code using Sionna. Due to the inherent stochasticity of LLMs, different runs of the same task produce different sets of algorithms with varying performance; however, the best-performing algorithms were found to be consistently competitive with conventional approaches.

After a run completes, the framework produces not a single algorithm but a ranked set of algorithms. In this report, we present results only for the best-performing algorithm of each run. The generated Python code is available as part of the released repository accompanying this paper. 
Detailed mathematical descriptions of the best-performing algorithms for each task and model are provided in the appendices.\footnote{These algorithm descriptions were drafted from the generated code by generative AI. We have edited and reformatted them only minimally.} 

\subsection{Channel Estimation}

The PHY layer tasks consist of generating a channel estimator for an orthogonal frequency-division multiplexing (OFDM) multiple-input multiple-output (MIMO) system. We consider two variants of the channel estimation problem, which define the two PHY layer tasks.

In the first variant, referred to as \emph{statistics-agnostic estimation}, no prior knowledge of the channel statistics is assumed. The agent must therefore generate algorithms that operate without access to any second-order channel information. However, the agents are free to make assumptions about the statistics.

In the second variant, referred to as \emph{estimation with known covariance}, the channel estimator is provided with the spatial, frequency, and temporal covariance matrices of the channel. During evaluation, the evaluation tool injects pre-computed covariance matrices for the channel model under test into the agent's workspace. The agent is instructed to generate algorithms that load and exploit these matrices.

In both cases, the channel model, number of transmitted streams, number of receive antennas, pilot pattern, number of subcarriers, and number of OFDM symbols are not disclosed to the agents, in order to encourage the generation of robust and general algorithms. The resulting code must therefore comply with 5G New Radio (NR) physical uplink shared channel (PUSCH) specifications and support arbitrary demodulation reference signal (DMRS) configurations, as well as arbitrary numbers of physical resource blocks (PRBs), transmitted streams, and OFDM symbols. It must also be written in PyTorch and be compatible with Sionna~2.0.

\subsubsection{Evaluation Setup}

The evaluation tool assesses the generated algorithms on an uplink 3GPP urban microcell (UMi) channel model, with four single-antenna user equipments (UEs) and a base station equipped with $16$ receive antennas. The OFDM resource grid consists of $72$ subcarriers and $14$ OFDM symbols, with reference signals placed on the 2nd and 11th symbols. Linear minimum mean square error (LMMSE) equalization and a~posteriori probability (APP) demapping are applied after channel estimation, which is not known to the agents.

As the framework requires the evaluation tool to return a single scalar metric, the normalized validation error (NVE)~\cite{gruber2017deep} is used. The NVE is defined as
\begin{equation}
\label{eq:nve}
\mathrm{NVE} = \frac{1}{N_{\mathrm{snr}}} \sum_{i=1}^{N_{\mathrm{snr}}} \frac{\mathrm{BLER}_{\mathrm{agent}}(\mathrm{SNR}_i)}{\mathrm{BLER}_{\mathrm{pCSI}}(\mathrm{SNR}_i)},
\end{equation}
where $\mathrm{BLER}_{\mathrm{agent}}$ and $\mathrm{BLER}_{\mathrm{pCSI}}$ denote the block error rates (BLERs) achieved by the agent's estimator and under perfect channel state information (CSI), respectively, and $\mathrm{SNR}_i$ denotes the $i$-th signal-to-noise ratio (SNR) evaluation point. The BLERs are estimated through Monte Carlo simulation using Sionna with GPU acceleration over an SNR range of $-9$ to $-2$~dB. The framework therefore aims to generate algorithms that minimize the NVE: an NVE equal to~$1$ corresponds to an estimator that achieves the same performance as perfect CSI. Note that an alternative metric would be the mean square error (MSE) between the agent's estimate and the true channel frequency response, but we chose the NVE to account for end-to-end detection performance, which includes equalization, demapping, and channel decoding.

\subsubsection{Results: Statistics-Agnostic Estimation}

\begin{figure}[t]
\centering
\begin{tikzpicture}

\begin{semilogyaxis}[standard_plotstyle, 
ymin=10, ymax=100, xmin=0, xmax=9, height=2.5cm,
ytick={10, 20, 30, 40, 50, 60, 70, 80, 90, 100},
yticklabels={10, , , , , , , , , 100},
xtick={0,...,9}, 
ylabel={best~\,NVE},
xlabel={iteration}
]
\addplot[nvidiapurple]
    table[col sep=comma, x=iteration, y=cumulative_best]{data/evolution_agnostic_gpt54.csv};

\addplot[nvidiagreen]
    table[col sep=comma, x=iteration, y=cumulative_best]{data/evolution_agnostic_gptoss.csv};

\node[nvidiapurple, labelstyle] at (5, 50) {GPT 5.4};
\node[nvidiagreen, labelstyle] at (4, 21) {GPT-OSS 120B};

\end{semilogyaxis}

\begin{axis}[standard_plotstyle, yshift=-4.cm, height=2.5cm,
xmin=0,
xmax=9,
ymin=0, 
ymax=1,
xtick={0,...,9},
yticklabels={, 0.0, 0.2, 0.4, 0.6, 0.8, 1.0},
ylabel={success rate},
xlabel={iteration},
enlarge y limits={upper, value=0.015}
]

\addplot[nvidiapurple]
    table[col sep=comma, x=iteration, y=success_rate]{data/evolution_agnostic_gpt54.csv};

\addplot[nvidiagreen]
    table[col sep=comma, x=iteration, y=success_rate]{data/evolution_agnostic_gptoss.csv};
    
\node[nvidiapurple, labelstyle] at (5.5, 0.9) {GPT 5.4};
\node[nvidiagreen, labelstyle] at (5, 0.5) {GPT-OSS 120B};

\end{axis}
\end{tikzpicture}
\caption{Evolution of the best NVE and success rate across iterations for the statistics-agnostic channel estimation task}
\label{fig:ce_evolution}
\end{figure}
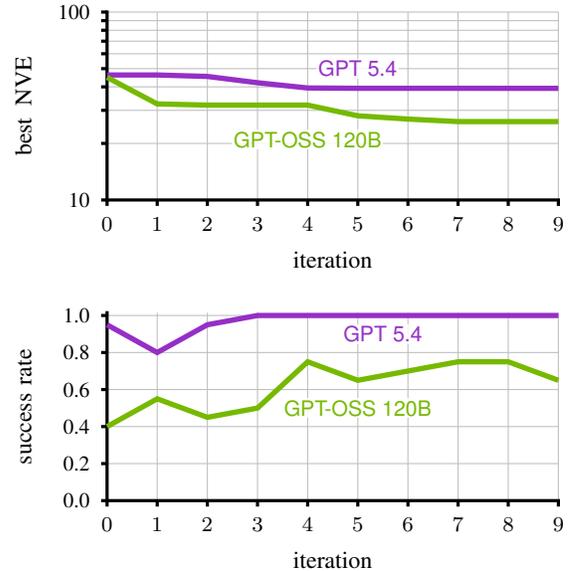

To provide some insight into the functioning of the framework, we start by presenting the evolution of the best NVE and success rate across iterations for the statistics-agnostic channel estimation task. Fig.~\ref{fig:ce_evolution} shows the lowest NVE achieved as a function of the iteration index when running the framework with GPT~5.4 and GPT-OSS~120B, together with the success rate, defined as the fraction of generated code that executes without crashing at each iteration. Running the framework over multiple iterations reduces the NVE, as the orchestrator refines ideas and the agents build upon algorithms from previous iterations. However, the success rate is notably higher with the more capable model GPT~5.4. Consequently, a run with GPT~5.4 produces a larger set of working algorithms compared to GPT-OSS~120B. Interestingly, despite the significantly lower success rate of GPT-OSS~120B, the best-performing algorithm generated with this model achieves a lower NVE than the one generated with GPT~5.4 on this task. A possible explanation for this performance gap is that, within the fixed time budget, GPT~5.4 agents performed approximately $22.5\%$ fewer turns, where a turn consists of an LLM call followed by a tool invocation, than GPT-OSS~120B agents, due to the higher per-call latency of the closed-weights model.

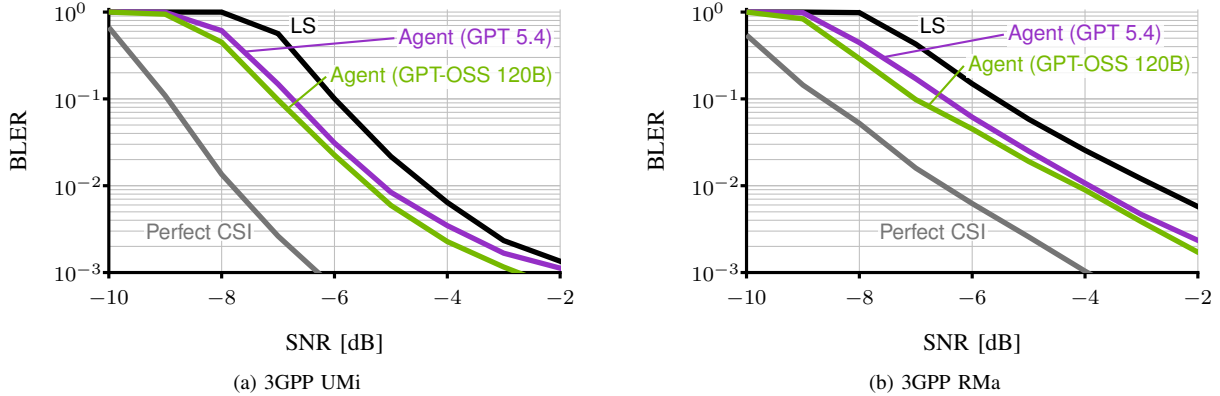
\begin{figure*}[t]
\centering
\subfloat[3GPP UMi]{%
\begin{tikzpicture}
\begin{semilogyaxis}[ber_plotstyle, 
xmin=-10, 
xmax=-2,
ymin=0.001,
ylabel=BLER,
enlarge y limits={upper, value=0.01}
]
\addplot[nvidiagray]
    table[col sep=comma, x=snr, y=perfect-csi]{data/bler_agnostic_umi.csv};
\node[nvidiagray, labelstyle] at (-8.4, 0.003) {Perfect CSI};

\addplot[black]
    table[col sep=comma, x=snr, y=baseline]{data/bler_agnostic_umi.csv};
\node[black, labelstyle] at (-6.55, 0.7) {LS};

\addplot[nvidiapurple]
    table[col sep=comma, x=snr, y=agent-gpt-5.4]{data/bler_agnostic_umi.csv};
\node[nvidiapurple, labelstyle] (x) at (-3.5, 0.5) {Agent (GPT 5.4)};
\draw[nvidiapurple, thick] (x.west) to (-7.6, 0.35);

\addplot[nvidiagreen]
    table[col sep=comma, x=snr, y=agent-gpt-oss-120b]{data/bler_agnostic_umi.csv};
\node[nvidiagreen, labelstyle] (y) at (-4.1, 0.18) {Agent (GPT-OSS 120B)};
\draw[nvidiagreen, thick] (y.west) to (-6.8, 0.08);

\end{semilogyaxis}
\end{tikzpicture}%
}
\hspace{0.5cm}
\subfloat[3GPP RMa]{%
\begin{tikzpicture}
\begin{semilogyaxis}[ber_plotstyle, 
xmin=-10, 
xmax=-2,
ymin=0.001,
ylabel=BLER,
enlarge y limits={upper, value=0.01}
]

\addplot[nvidiagray]
    table[col sep=comma, x=snr, y=perfect-csi]{data/bler_agnostic_rma.csv};
\node[nvidiagray, labelstyle] at (-6.7, 0.003) {Perfect CSI};

\addplot[black]
    table[col sep=comma, x=snr, y=baseline]{data/bler_agnostic_rma.csv};
\node[black, labelstyle] at (-6.7, 0.7) {LS};

\addplot[nvidiapurple]
    table[col sep=comma, x=snr, y=agent-gpt-5.4]{data/bler_agnostic_rma.csv};
\node[nvidiapurple, labelstyle] (x) at (-4., 0.55) {Agent (GPT 5.4)};
\draw[nvidiapurple, thick] (x.west) to (-7.6, 0.3);

\addplot[nvidiagreen]
    table[col sep=comma, x=snr, y=agent-gpt-oss-120b]{data/bler_agnostic_rma.csv};
\node[nvidiagreen, labelstyle] (y) at (-4.1, 0.24) {Agent (GPT-OSS 120B)};
\draw[nvidiagreen, thick] (y.west) to (-6.8, 0.08);
\end{semilogyaxis}
\end{tikzpicture}%
}
\caption{BLER curves for the statistics-agnostic channel estimation task}
\label{fig:ce_bler_nocov}
\end{figure*}

We now examine the end-to-end detection performance of the best-performing algorithms. Fig.~\ref{fig:ce_bler_nocov} shows the BLER curves achieved by the best-performing algorithms generated with GPT~5.4 and GPT-OSS~120B, alongside two baselines: perfect CSI and least squares (LS) estimation at pilot positions with linear interpolation. Results are shown for the 3GPP UMi channel used during evaluation, as well as for a 3GPP rural macrocell (RMa) channel that was \emph{not} seen during evaluation, in order to assess generalization. Both algorithms generated by the framework significantly outperform the LS baseline. The algorithm derived with GPT-OSS~120B achieves lower error rates than the one from GPT~5.4 at low SNR, demonstrating that open-weights models can be competitive for algorithm generation.

To gain insight into the generated algorithms, we provide an overview of their functioning.
Since the generated code is standard Python, it can be readily inspected, making the resulting algorithms fully explainable and extensible, in contrast to neural network-based approaches.
The full code is available in the repository accompanying this paper.
Interestingly, the two best-performing algorithms adopt markedly different approaches, illustrating the framework's ability to generate diverse, non-trivial signal processing algorithms.
The best-performing algorithm generated with GPT-OSS~120B (see Appendix~\ref{sec:app_ce_agnostic_gptoss}) refines an initial LS channel estimate by applying orthogonal matching pursuit (OMP) at pilot locations to denoise the estimate through time-domain sparsity enforcement. It then estimates the Doppler correlation and constructs an adaptive time-domain finite impulse response (FIR) filter. A forward Kalman filter combined with a backward Rauch--Tung--Striebel (RTS) smoother tracks the channel evolution across all OFDM symbols. Finally, a second FIR filter and a frequency-domain B-spline filter smooth the reconstructed channel along both time and frequency, yielding a refined channel estimate and its corresponding error variance.

The best-performing algorithm generated with GPT~5.4 (see Appendix~\ref{sec:app_ce_agnostic_gpt54}) implements an iterative graph-based message-passing channel estimator on the two-dimensional time-frequency resource grid. It treats the grid as a connected graph in which pilot symbols act as anchored observation nodes. The algorithm first creates an initial coarse estimate using a weighted combination of the nearest pilots, then refines it over several iterations by averaging neighboring nodes with dynamic edge weights derived from the phase mismatch between adjacent pilots. It assigns a spatially varying error variance to each node based on its composite distance to the nearest pilots, local pilot quality, and convergence stability.

\subsubsection{Results: Estimation with Known Covariance}

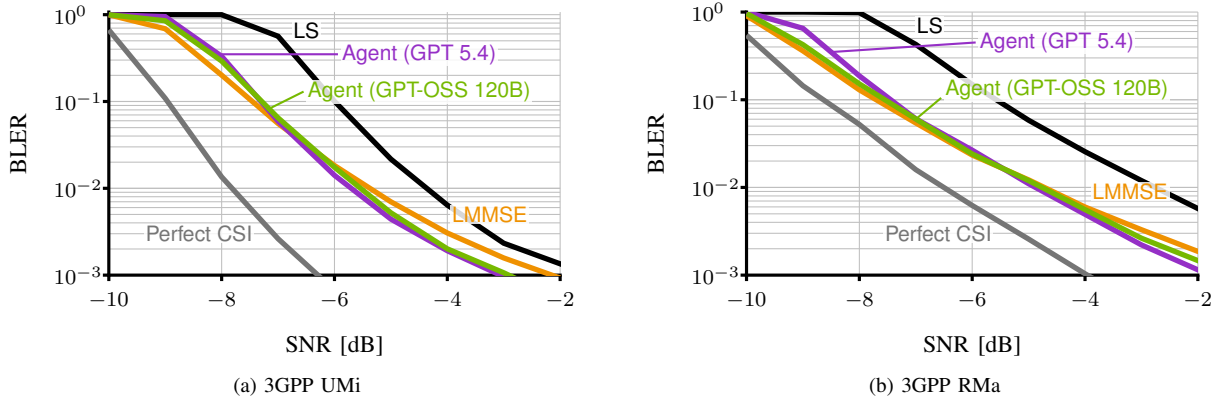
\begin{figure*}[t]
\centering
\subfloat[3GPP UMi]{%
\begin{tikzpicture}
\begin{semilogyaxis}[ber_plotstyle, 
xmin=-10, 
xmax=-2,
ymin=0.001,
ylabel=BLER,
enlarge y limits={upper, value=0.01},
]

\addplot[nvidiagray]
    table[col sep=comma, x=snr, y=perfect-csi]{data/bler_cov_umi.csv};
\node[nvidiagray, labelstyle] at (-8.4, 0.003) {Perfect CSI};

\addplot[black]
    table[col sep=comma, x=snr, y=baseline]{data/bler_cov_umi.csv};
\node[black, labelstyle] at (-6.5, 0.65) {LS};

\addplot[nvidiaorange]
    table[col sep=comma, x=snr, y=baseline-lmmse]{data/bler_cov_umi.csv};
\node[nvidiaorange, labelstyle] at (-3.25, 0.005) {LMMSE};

\addplot[nvidiapurple]
    table[col sep=comma, x=snr, y=agent-gpt-5.4]{data/bler_cov_umi.csv};
\node[nvidiapurple, labelstyle] (x) at (-4.5, 0.35) {Agent (GPT 5.4)};
\draw[nvidiapurple, thick] (x.west) to (-8, 0.35);

\addplot[nvidiagreen]
    table[col sep=comma, x=snr, y=agent-gpt-oss-120b]{data/bler_cov_umi.csv};
\node[nvidiagreen, labelstyle] (y) at (-4.5, 0.13) {Agent (GPT-OSS 120B)};
\draw[nvidiagreen, thick] (y.west) to (-7.2, 0.08);

\end{semilogyaxis}
\end{tikzpicture}%
}
\hspace{0.5cm}
\subfloat[3GPP RMa]{%
\begin{tikzpicture}
\begin{semilogyaxis}[ber_plotstyle, 
xmin=-10, 
xmax=-2,
ymin=0.001,
ylabel=BLER
]

\addplot[nvidiagray]
    table[col sep=comma, x=snr, y=perfect-csi]{data/bler_cov_rma.csv};
\node[nvidiagray, labelstyle] at (-6.6, 0.003) {Perfect CSI};

\addplot[black]
    table[col sep=comma, x=snr, y=baseline]{data/bler_cov_rma.csv};
\node[black, labelstyle] at (-6.75, 0.7) {LS};

\addplot[nvidiaorange]
    table[col sep=comma, x=snr, y=baseline-lmmse]{data/bler_cov_rma.csv};
\node[nvidiaorange, labelstyle] at (-3.2, 0.009) {LMMSE};

\addplot[nvidiapurple]
    table[col sep=comma, x=snr, y=agent-gpt-5.4]{data/bler_cov_rma.csv};
\node[nvidiapurple, labelstyle] (x) at (-4.5, 0.45) {Agent (GPT 5.4)};
\draw[nvidiapurple, thick] (x.west) to (-8.5, 0.35);

\addplot[nvidiagreen]
    table[col sep=comma, x=snr, y=agent-gpt-oss-120b]{data/bler_cov_rma.csv};
\node[nvidiagreen, labelstyle] (y) at (-4.5, 0.13) {Agent (GPT-OSS 120B)};
\draw[nvidiagreen, thick] (-6.55,0.105) to (-7., 0.06);

\end{semilogyaxis}
\end{tikzpicture}%
}
\caption{BLER curves for channel estimation with known covariance}
\label{fig:ce_bler_cov}
\end{figure*}

We now turn to the second channel estimation variant, in which agents are provided with the time, frequency, and spatial covariance matrices of the channel. Fig.~\ref{fig:ce_bler_cov} shows the corresponding BLER curves. As an additional baseline, we include LS estimation with LMMSE interpolation, which applies LMMSE filtering along the time, frequency, and spatial dimensions and therefore also requires knowledge of the corresponding covariance matrices.

For this task, the algorithms generated by GPT-OSS~120B and GPT~5.4 achieve similar error rates, on par with those of the LMMSE baseline. A closer inspection of the generated algorithms reveals them to be specialized variations of the LMMSE filter, confirming the framework's ability to generate non-trivial algorithms but also highlighting a limitation: when a well-known strong solution exists (LMMSE under known covariance), the framework may converge toward a variant of it rather than discovering fundamentally novel approaches.

The algorithm generated with GPT-OSS~120B (see Appendix~\ref{sec:app_ce_cov_gptoss}) implements an iterative Kronecker-structured LMMSE estimator. Since the covariance matrices along the time, frequency, and spatial dimensions are provided, their eigendecompositions can be pre-computed. The algorithm projects the noisy LS estimate into the resulting uncorrelated eigendomain and applies a two-pass adaptive Wiener filter iteratively, with a damping factor ensuring stable convergence.

The algorithm generated with GPT~5.4 (see Appendix~\ref{sec:app_ce_cov_gpt54}) implements a separable, sequential LMMSE estimator. It dynamically generates regularized filtering matrices for each dimension based on the current noise variance, sequentially applies them as linear transformations along their respective axes, and constructs the final estimate as a weighted combination of the initial linearly interpolated LS estimate and the intermediate filtered states.

\subsection{Link Adaptation}

The MAC layer task consists in implementing a modulation and coding scheme (MCS) selection controller for link adaptation. The controller must maximize spectral efficiency (SE) while maintaining the BLER below a predefined target of $10\%$. It observes a series of hybrid automatic repeat request (HARQ) acknowledgement/negative acknowledgement (ACK/NACK) feedback together with the corresponding MCS indices, and selects the MCS index for the next transmission. Crucially, the controller has no direct access to the underlying signal-to-interference-plus-noise ratio (SNR) values.

\subsubsection{Evaluation Setup}

The evaluation tool assesses a candidate MCS controller by replaying $50$~pre-generated SNR trajectories through a simulated link adaptation loop. The SNR trajectories were generated by simulating UE mobility within an urban scene (Munich) using Sionna~RT~\cite{sionna-rt}. For each trajectory, a UE moves along a straight path at a random speed between $3$ and $14$~m/s, and the channel frequency response is computed at every slot ($0.5$~ms) via ray tracing. The length of the trajectories was set to $3000$~slots, corresponding to $1.5$~seconds of simulated time. The per-slot channel gain is obtained by averaging the squared magnitude of the channel frequency response over all subcarriers, yielding the effective SNR per slot.

During evaluation, the simulation replays each SNR trajectory slot by slot. At each slot, the BLER for the current (SNR, MCS) pair is computed using pre-fitted BLER-vs-SNR curves, and a Bernoulli draw determines whether the transmission results in an ACK or a NACK. HARQ feedback is delivered to the MCS selection algorithm in batches of $5$~slots to emulate feedback delay. The evaluation tool computes two metrics per scenario: the long-term BLER, which must remain below the target, and the average SE, the optimization objective. The final metric reported to the agent is the SE averaged across all $50$~scenarios, and the evaluation is marked as successful only if the BLER constraint is satisfied on every scenario. As additional information, the tool reports BLER statistics across all trajectories, as well as the number of trajectories that did not meet the BLER constraint.

\subsubsection{Results}

As baseline, we consider outer loop link adaptation (OLLA)~\cite{pedersen2007frequency} for MCS selection. For fairness, the baseline hyperparameter that controls the SNR back-off step size applied after each failed transmission was fine-tuned on the trajectories used by the evaluation tool to maximize the SE while satisfying the BLER constraint on all sequences. Table~\ref{tab:la_results} shows the benchmarking results, computed on a separate non-overlapping set of $50$~trajectories also generated with Sionna~RT on the Munich scene, to ensure that the selected algorithms have not overfit to the evaluation set. All approaches satisfy the BLER constraint on all $50$~sequences. The algorithm generated with GPT-OSS~120B achieves an SE slightly lower than the fine-tuned OLLA baseline, while the algorithm generated with GPT~5.4 achieves a significant SE gain of more than~$3\%$, showing that agentic AI can generate MAC layer algorithms that are competitive with conventional approaches.

\begin{table}[t]
    \centering
    \caption{Link adaptation benchmarking results on the held-out set. SE denotes spectral efficiency.}
    \label{tab:la_results}
    \begin{tabular}{lccc}
        \toprule
        Algorithm & Avg.\ SE (bps/Hz) & Success & vs.\ Baseline \\
        \midrule
        GPT-OSS 120B  & 3.3976 & 50/50 & $-0.64\%$ \\
        GPT 5.4       & 3.5370 & 50/50 & $+3.43\%$ \\
        OLLA baseline & 3.4196 & 50/50 & --- \\
        \bottomrule
    \end{tabular}
\end{table}

To gain insight into the generated link adaptation algorithms, we provide an overview of their functioning.
The best-performing algorithm generated with GPT-OSS~120B (see Appendix~\ref{sec:app_la_gptoss}) implements a Monte Carlo look-ahead link adaptation strategy based on a one-dimensional particle filter. It tracks the posterior distribution of the unobservable SNR using a set of weighted particles that evolve via a random walk and are updated using binary ACK/NACK feedback. To select the MCS, it evaluates the expected immediate and one-step-ahead SE for all candidate MCS values across the particle distribution.

The best-performing algorithm generated with GPT~5.4 (see Appendix~\ref{sec:app_la_gpt54}) is simpler. It implements a grid-based discrete Markov filter paired with an asymmetric OLLA mechanism. It discretizes the SNR space into a fixed grid and tracks the probability distribution of the true SNR over time using Gaussian convolution for time evolution and Bayesian updates based on ACK/NACK feedback. A trimmed mean of the posterior distribution yields an SNR estimate, which is then adjusted by an asymmetric OLLA offset, stepping up for ACKs and stepping down more aggressively for NACKs, before selecting the final MCS.

\section{Conclusion}
\label{sec:conclusion}

This paper explores the ability of agentic AI to autonomously design wireless communication algorithms. To that end, we built a dedicated framework and evaluated it on three tasks spanning both the PHY and MAC layers. Using off-the-shelf LLMs, the framework generates competitive algorithms in a matter of hours: the best-performing algorithms significantly outperform simple baselines, match strong classical methods such as LMMSE channel estimation, and in the case of link adaptation, surpass a fine-tuned OLLA baseline by more than~$3\%$ in spectral efficiency.

A noteworthy property of the agentic approach is that the generated algorithms consist of conventional signal processing code, fully explainable and extensible. This stands in contrast to neural network methods whose internal workings are difficult to interpret.

However, these early experiments also reveal a limitation of current agentic AI systems: when a strong classical solution exists, the framework tends to converge toward variations of it rather than discovering fundamentally novel approaches. It remains open for future work to investigate whether this lack of creativity is an inherent limitation of LLM-based research or whether our choice of experimental problems did not allow for significant improvements over the state of the art. We note that the framework is at an early stage of development, and many improvements are possible. Advancing agentic AI to enable the discovery of genuinely novel wireless communication algorithms is a promising and challenging direction, as this form of algorithmic design may become common practice.

\bibliographystyle{IEEEtran}
\bibliography{bibliography}

\appendices

\section{Statistics-Agnostic Channel Estimation: Generated Algorithms}

\subsection{GPT-OSS~120B}
\label{sec:app_ce_agnostic_gptoss}
This algorithm implements a channel estimation and tracking pipeline for OFDM-based systems. It refines an initial noisy LS channel estimate through a sequence of signal processing techniques designed to exploit both the sparse nature of multipath channels and their temporal-frequency correlations. The core functioning consists of applying OMP at pilot locations to denoise the estimates by enforcing time-domain sparsity. It then estimates the Doppler correlation to generate an adaptive time-domain FIR filter. A forward Kalman filter combined with a backward RTS smoother tracks the channel evolution across all OFDM symbols (both pilots and data). Finally, a second time-domain FIR filter and a frequency-domain B-spline filter are applied to smooth the reconstructed channel across both time and frequency axes, yielding a highly refined channel estimate and its corresponding error variance.

Let $M$ be the number of subcarriers (FFT size) and $T$ be the total number of OFDM symbols. The algorithm relies on the following predefined hyperparameters: $K = 4$ for the OMP sparsity level, $L_{\text{max}} = 16$ for the maximum channel tap delay, and $L_{\text{FIR}} = 7$ for the length of the time-domain FIR kernel.

\subsubsection{Initial LS Estimation}
The algorithm begins with a baseline LS channel estimator. Let the initial channel estimates over time $t$ and frequency $f$ be $\hat{H}_{\text{LS}}(t, f)$ and the corresponding error variance be $\sigma^2_{\text{LS}}(t, f)$. Let $\mathcal{P}$ denote the set of time indices $t$ that contain pilot symbols.

\subsubsection{OMP Sparsification}
For the time indices $t \in \mathcal{P}$, the algorithm suppresses noise by selecting the $K$ most dominant time-domain taps.
\begin{itemize}
    \item \textrm{Dictionary Construction:} A discrete Fourier transform (DFT) dictionary $A \in \mathbb{C}^{M \times L_{\text{max}}}$ is built. The elements are defined as
    \[ A_{m, l} = \exp\left(-j \frac{2\pi m l}{M}\right), \]
    where $m \in \{0, \dots, M-1\}$ and $l \in \{0, \dots, L_{\text{max}}-1\}$.
    \item \textrm{Correlation \& Support Selection:} The correlation $c(t, l)$ is computed
    \[ c(t, l) = \sum_{m=0}^{M-1} \hat{H}_{\text{LS}}(t, m) A_{m, l}^*. \]
    The support set $\Omega(t)$ of size $K$ is formed by taking the indices $l$ that yield the largest magnitude $|c(t, l)|$.
    \item \textrm{Least Squares on Support:} Let $A_{\Omega(t)} \in \mathbb{C}^{M \times K}$ be the sub-matrix of $A$. The sparse time-domain taps $\hat{x}(t) \in \mathbb{C}^K$ are estimated via
    \[ \hat{x}(t) = (A_{\Omega(t)}^H A_{\Omega(t)})^{-1} A_{\Omega(t)}^H \hat{H}_{\text{LS}}(t). \]
    \item \textrm{Reconstruction:}
    \begin{align*}\hat{H}_{\text{OMP}}(t) &= A_{\Omega(t)} \hat{x}(t) \\
     E_{\text{OMP}}(t, f) &= |\hat{H}_{\text{LS}}(t, f) - \hat{H}_{\text{OMP}}(t, f)|^2
     \end{align*}
\end{itemize}

\subsubsection{Doppler Coefficient Estimation}
The temporal correlation $a_{\text{est}}$ is calculated using the OMP estimates
\begin{align*}
a_{\text{est}} = \text{clamp}\biggl(
  \frac{\mathbb{E}_{t,f}[\Re(\hat{H}_{\text{OMP}}(t, f)\, \hat{H}_{\text{OMP}}^*(t\!+\!1, f))]}
       {\mathbb{E}_{t,f}[|\hat{H}_{\text{OMP}}(t, f)|^2] + \epsilon},
  0.5, 0.99 \biggr)
\end{align*}
where $\epsilon = 10^{-12}$ is a small constant added to the denominator to prevent division by zero, and $\text{clamp}(x, a, b)$ clamps $x$ to the interval $[a, b]$.

\subsubsection{Time-Domain FIR Filtering (Pilots)}
A symmetric FIR kernel $w_t(\tau)$ of length $L_{\text{FIR}}$ is constructed as
\[ w_t(\tau) = \frac{a_{\text{est}}^{|\tau|}}{\sum_{i} a_{\text{est}}^{|i|}}. \]
This filter is applied via convolution over the time dimension,
\begin{align*}
\tilde{H}_{\text{pilot}}(t, f) &= (\hat{H}_{\text{OMP}} * w_t)(t, f), \\
\tilde{E}_{\text{pilot}}(t, f) &= (E_{\text{OMP}} * w_t)(t, f).
\end{align*}

\subsubsection{Forward Kalman Filter}
A scalar Kalman filter tracks the channel across all OFDM symbols $t \in \{0, \dots, T-1\}$. Let $a_k$ be the state transition coefficient of a first-order Autoregressive (AR(1)) model used as a generic prior (set to $0.99$), and let $q = 1 - a_k^2$ denote the corresponding process noise variance.
\begin{itemize}
    \item \textrm{Initialization ($t=0$):} If $0 \in \mathcal{P}$, $\hat{H}_{\text{fwd}}(0) = \tilde{H}_{\text{pilot}}(0)$ and $P_{\text{fwd}}(0) = \tilde{E}_{\text{pilot}}(0)$. Otherwise, initialized with $\hat{H}_{\text{LS}}(0)$ and $\sigma^2_{\text{LS}}(0)$.
    \item \textrm{Prediction Step ($t > 0$):}
    \begin{align*}
    H_{\text{pred}}(t) &= a_k \hat{H}_{\text{fwd}}(t-1), \\
    P_{\text{pred}}(t) &= a_k^2 P_{\text{fwd}}(t-1) + q 
    \end{align*}
    \item \textrm{Measurement Update Step ($t \in \mathcal{P}$):}
    \begin{align*}
    K_{\text{gain}}(t) &= \frac{P_{\text{pred}}(t)}{P_{\text{pred}}(t) + \tilde{E}_{\text{pilot}}(t)}, \\
    \hat{H}_{\text{fwd}}(t) &= H_{\text{pred}}(t) + K_{\text{gain}}(t) \left( \tilde{H}_{\text{pilot}}(t) - H_{\text{pred}}(t) \right), \\
    P_{\text{fwd}}(t) &= (1 - K_{\text{gain}}(t)) P_{\text{pred}}(t).
    \end{align*}
    If $t \notin \mathcal{P}$, $\hat{H}_{\text{fwd}}(t) = H_{\text{pred}}(t)$ and $P_{\text{fwd}}(t) = P_{\text{pred}}(t)$.
\end{itemize}

\subsubsection{Backward RTS Smoother}
\begin{itemize}
    \item \textrm{Initialization:} 
    \begin{align*}
    \hat{H}_{\text{sm}}(T-1) &= \hat{H}_{\text{fwd}}(T-1), \\
    P_{\text{sm}}(T-1) &= P_{\text{fwd}}(T-1).
    \end{align*}
    \item \textrm{Smoothing Step:}
    \begin{align*}
    C(t) &= \frac{P_{\text{fwd}}(t) a_k}{P_{\text{pred}}(t+1)}, \\
    \hat{H}_{\text{sm}}(t) &= \hat{H}_{\text{fwd}}(t) + C(t) \left( \hat{H}_{\text{sm}}(t+1) - H_{\text{pred}}(t+1) \right), \\
    P_{\text{sm}}(t) &= P_{\text{fwd}}(t) + C(t)^2 \left( P_{\text{sm}}(t+1) - P_{\text{pred}}(t+1) \right).
    \end{align*}
\end{itemize}

\subsubsection{Final Time and Frequency Smoothing}
\begin{itemize}
    \item \textrm{Time FIR Filter:} The adaptive kernel $w_t(\tau)$ is applied across the entire time axis $t$,
    \[ \hat{H}_{\text{time}}(t, f) = (\hat{H}_{\text{sm}} * w_t)(t, f), \]
    \[ P_{\text{time}}(t, f) = (P_{\text{sm}} * w_t)(t, f). \]
    \item \textrm{Frequency Spline Filter:} A fixed 1D discrete convolution kernel approximating a cubic B-spline, defined as $w_f = [0.125, 0.375, 0.5, 0.375, 0.125]$, is applied across the frequency axis $f$,
    \[ \hat{H}_{\text{final}}(t, f) = (\hat{H}_{\text{time}} * w_f)(t, f), \]
    \[ P_{\text{final}}(t, f) = (P_{\text{time}} * w_f)(t, f). \]
\end{itemize}

\subsection{GPT~5.4}
\label{sec:app_ce_agnostic_gpt54}
This algorithm implements an iterative, graph-based message-passing channel estimator on the 2D time-frequency resource grid. It interpolates missing channel estimates by treating the resource grid as a connected graph where pilot symbols act as anchored observation nodes. The algorithm initializes a coarse estimate using a weighted combination of the nearest pilots. It then refines this state over several iterations by averaging neighboring nodes, utilizing dynamic edge weights derived from the phase mismatch between adjacent pilots. Finally, it assigns a spatially varying error variance to each node based on its composite distance to the nearest pilots, local pilot quality, and solver convergence stability.

Let $N_{\text{t}}$ be the number of OFDM symbols (time dimension) and $N_{\text{f}}$ be the number of subcarriers (frequency dimension). Let $(t, f)$ denote the discrete coordinates on this grid, where $t \in \{0, \dots, N_{\text{t}}-1\}$ and $f \in \{0, \dots, N_{\text{f}}-1\}$.

Let $\hat{H}_{\text{LS}}(t, f)$ be the initial LS channel estimate, $\sigma^2_{\text{LS}}(t, f)$ be the corresponding LS error variance, and $N_0$ be the scalar noise variance. Let $M(t, f) \in \{0, 1\}$ be the binary pilot mask, where $1$ indicates a pilot location and $0$ indicates a data location. 

\subsubsection{Distance Mapping and Pilot Support Density}
The algorithm computes spatial metrics relative to the pilot positions to dictate regional smoothing behavior.
\begin{itemize}
    \item \textrm{Distances:} Let $d_{\text{t}}(t, f)$ and $d_{\text{f}}(t, f)$ be the absolute distance (in indices) from coordinate $(t, f)$ to the nearest pilot in the time and frequency axes, respectively. A composite distance score $d_{\text{score}}(t, f)$ is defined as
    \[ d_{\text{score}}(t, f) = d_{\text{f}}(t, f) + 1.55\, d_{\text{t}}(t, f). \]
    \item \textrm{Region Partitioning:} Non-pilot elements ($M(t,f)=0$) are partitioned into three mutually exclusive regions based on $d_{\text{score}}(t, f)$:
    \begin{itemize}
        \item Near Region: $d_{\text{score}}(t, f) \le 1.7$
        \item Mid Region: $1.7 < d_{\text{score}}(t, f) \le 3.55$
        \item Far Region: $d_{\text{score}}(t, f) > 3.55$
    \end{itemize}
    \item \textrm{Pilot Support Density ($S$):} The local pilot density $S(t, f)$ is initialized as the float representation of the mask: $S^{(0)}(t, f) = M(t, f)$. It is updated over $k \in \{1, 2\}$ iterations via a spatial blur,
    \begin{align*}
    S^{(k)}(t, f) &= 0.30\, S^{(k-1)}(t, f) \\
      &+ 0.27 \bigl(S^{(k-1)}(t, f\!-\!1) + S^{(k-1)}(t, f\!+\!1)\bigr) \\
      &+ 0.08 \bigl(S^{(k-1)}(t\!-\!1, f) + S^{(k-1)}(t\!+\!1, f)\bigr).
    \end{align*}
    The final support is clamped,
    \[S(t, f) = \max(S^{(2)}(t, f), 10^{-3}).\]
\end{itemize}

\subsubsection{Initial State Construction}
A starting channel state $H^{(0)}(t, f)$ is generated for the iterative solver.
\begin{itemize}
    \item Let $H_{\text{t}}(t,f)$ and $H_{\text{f}}(t,f)$ be the LS channel values of the temporally and frequency nearest pilots to $(t, f)$, respectively.
    \item A base blend is initialized: 
    \[ H_{\text{base}}^{(0)}(t, f) = 0.66\, H_{\text{f}}(t, f) + 0.34\, H_{\text{t}}(t, f) \]
    \item The base blend undergoes $k \in \{1, 2\}$ passes of a cross-shaped smoothing filter:
    \begin{multline*}
    H_{\text{base}}^{(k)}(t, f) = 0.28\, H_{\text{base}}^{(k-1)}(t, f) \\
      + 0.31 \bigl(H_{\text{base}}^{(k-1)}(t, f\!-\!1) + H_{\text{base}}^{(k-1)}(t, f\!+\!1)\bigr) \\
      + 0.05 \bigl(H_{\text{base}}^{(k-1)}(t\!-\!1, f) + H_{\text{base}}^{(k-1)}(t\!+\!1, f)\bigr)
    \end{multline*}
    \item \textrm{Pilot Anchor:} The initial state $H^{(0)}(t, f)$ strictly anchors pilot locations to their true LS values:
    \[ H^{(0)}(t, f) = \begin{cases} \hat{H}_{\text{LS}}(t, f), & M(t, f) = 1 \\ H_{\text{base}}^{(2)}(t, f), & M(t, f) = 0 \end{cases} \]
\end{itemize}

\subsubsection{Edge Weight Computation (Phase Mismatch)}
Dynamic transition weights (gates) are computed to preserve channel edges. Let the function $\text{clamp}(x, a, b)$ denote restricting the value of $x$ to the interval $[a, b]$.
\begin{itemize}
    \item \textrm{Phase Mismatch Function:} For any two adjacent complex estimates $a$ and $b$, the phase correlation is $c = a b^*$. The unit phase vector is $\phi = \frac{c}{\max(|c|, 10^{-4})}$. The mismatch $m(a, b)$ is:
    \[ m(a, b) = \frac{|a - \phi b|}{0.5 \max(|a| + |b|,\; 2 \cdot 10^{-3})} \]
    \item Let $m_{\text{t}}(t, f) = m(\hat{H}_{\text{LS}}(t, f), \hat{H}_{\text{LS}}(t-1, f))$ and $m_{\text{f}}(t, f) = m(\hat{H}_{\text{LS}}(t, f), \hat{H}_{\text{LS}}(t, f-1))$ be the mismatches. These are strictly evaluated only if both nodes in the pair are valid pilots; otherwise, $m_{\text{t}}(t, f) = 0$ and $m_{\text{f}}(t, f) = 0$.
    \item \textrm{Transmission Gates:} The directional gates $g_{\text{t}}(t, f)$ and $g_{\text{f}}(t, f)$ are defined as:
    \begin{align*}
    g_{\text{t}}(t, f) &= \text{clamp}(1.0 - 0.08\, m_{\text{t}}(t, f),\; 0.92,\; 0.985) \\
    g_{\text{f}}(t, f) &= \text{clamp}(1.0 - 0.14\, m_{\text{f}}(t, f),\; 0.94,\; 0.997)
    \end{align*}
\end{itemize}

\subsubsection{Iterative Graph Smoothing}
The channel $H^{(i)}(t, f)$ is updated as follows (for $i \in \{0, \dots, 6\}$):
\begin{itemize}
    \item \textrm{Static Weights:} The observation and directional transition weights are constant across iterations,
    \begin{align*}
    W_{\text{obs}}(t, f) &= M(t, f)\\
      &~~\cdot \text{clamp}\!\left(0.16 + \frac{0.56}{\max(N_0, 10^{-4})},\; 0,\; 8.0\right) \\
    W_L &= 1.235 \cdot g_{\text{f}}(t, f), \\
    W_R &= 1.235 \cdot g_{\text{f}}(t, f+1), \\
    W_U &= 0.055 \cdot g_{\text{t}}(t, f), \\
    W_D &= 0.055 \cdot g_{\text{t}}(t+1, f)
    \end{align*}
    \item \textrm{Update Rule:} At each iteration $i$, a new unblended state $H_{\text{new}}(t, f)$ is computed:
    \begin{multline*}
    H_{\text{new}} = \frac{W_{\text{obs}} \hat{H}_{\text{LS}} + W_L H^{(i)}_{\text{left}} + W_R H^{(i)}_{\text{right}}}{W_{\text{obs}} + W_L + W_R + W_U + W_D + 10^{-6}}\\
      + \frac{W_U H^{(i)}_{\text{up}} + W_D H^{(i)}_{\text{down}}}{W_{\text{obs}} + W_L + W_R + W_U + W_D + 10^{-6}},
    \end{multline*}
    where 
    \begin{align*}
    H^{(i)}_{\text{left}} &= H^{(i)}(t, f-1), \\ 
    H^{(i)}_{\text{right}} &= H^{(i)}(t, f+1),\\ 
    H^{(i)}_{\text{up}} &= H^{(i)}(t-1, f), \\ 
    H^{(i)}_{\text{down}} &= H^{(i)}(t+1, f).
    \end{align*}
    \item Let the momentum factor be $\beta^{(i)} = 0.972$ for $i < 5$ and $\beta^{(i)} = 0.982$ for $i \ge 5$. The state is updated via a momentum blend and re-anchored to the pilots:
    \[ H^{(i+1)}\! = \!\begin{cases} \hat{H}_{\text{LS}}(t, f), & M = 1 \\ \beta^{(i)} H_{\text{new}} + (1 \!-\! \beta^{(i)}) H^{(i)}, & M = 0 \end{cases}, \]
    where the $(t, f)$ arguments are omitted for brevity.
\end{itemize}

\subsubsection{Final Frequency Smoothing}
After 7 iterations, a localized smoothing pass is applied strictly along the frequency axis to the final state $H^{(7)}(t, f)$:
\begin{align*}
H_{\text{freq}}(t, f) &= 0.16\, H^{(7)}(t, f) \\
  &~~+ 0.42\, H^{(7)}(t, f\!-\!1) + 0.42\, H^{(7)}(t, f\!+\!1).
\end{align*}
The output channel estimate $\hat{H}_{final}(t, f)$ is $H_{\text{freq}}(t, f)$ for data nodes and $\hat{H}_{\text{LS}}(t, f)$ for pilot nodes.

\subsubsection{Error Variance Assignment}
The final error variance $E_{\text{var}}(t, f)$ is constructed regionally based on local metrics:
\begin{itemize}
    \item Let $\text{MSE}_{\text{p}}$ be the scalar mean of $|\hat{H}_{\text{LS}}(t,f) - H^{(7)}(t,f)|^2$ evaluated strictly over all pilot nodes. Let $\bar{e}_0$ be the scalar global mean of the LS variance $\sigma^2_{\text{LS}}(t, f)$. Let $H^{(6)}(t, f)$ and $H^{(7)}(t, f)$ denote the channel states immediately before and after the final (seventh) message-passing iteration, respectively.
    \item \textrm{Local Metrics:}
    \begin{multline*}
    C(t, f) = \text{clamp}\!\left( \frac{|H^{(7)}(t, f) - H^{(6)}(t, f)|}{\max(|H^{(7)}(t, f)|, 3 \!\cdot\! 10^{-3})},\right.\\
        \left. 0.0,\; 0.5 \right)
    \end{multline*}
    \[ S_{\text{exc}}(t, f) = \max\!\left( \frac{1}{\max(S(t, f), 0.28)} - 1,\; 0 \right) \]
    \[ Q_{\text{p}} = \text{clamp}\!\left( \frac{\text{MSE}_{\text{p}}}{0.25 \bar{e}_0 + 10^{-4}},\; 0,\; 4 \right) \]
    \item \textrm{Distance Penalties:}
    \begin{align*}
    P_{\text{near}} &= 0.0013 + 0.0028\, d_{\text{score}} + 0.0018\, S_{\text{exc}} \\
    P_{\text{mid}}  &= 0.0048 + 0.0055\, d_{\text{score}} + 0.0042\, S_{\text{exc}} \\
    P_{\text{far}}  &= 0.0092 + 0.0080\, d_{\text{score}} + 0.0070\, S_{\text{exc}}\\
             &\quad + 0.0038\, Q_{\text{p}}
    \end{align*}
    \item \textrm{Distance Penalties:}
    \begin{align*}
    P_{\text{near}} &= 0.0013 + 0.0028\, d_{\text{score}} + 0.0018\, S_{\text{exc}} \\
    P_{\text{mid}}  &= 0.0048 + 0.0055\, d_{\text{score}} + 0.0042\, S_{\text{exc}} \\
    P_{\text{far}}  &= 0.0092 + 0.0080\, d_{\text{score}} + 0.0070\, S_{\text{exc}} + 0.0038\, Q_{\text{p}}
    \end{align*}
    \item \textrm{Final Variance Equations:} For the nodes for which \mbox{$M(t, f) = 1$}, the equations are
    \[ E_{\text{var}} = \max(0.26\, \sigma^2_{\text{LS}} + 0.006\, N_0,\; 10^{-4}), \]
    and for non-pilot nodes, depending on their region, the equations are
    \begin{align*}
        E_{\text{near}} &= 0.070\, \sigma^2_{\text{LS}} + 0.060\, \text{MSE}_{\text{p}} + P_{\text{near}} + 0.0095\, C, \\
        E_{\text{mid}}  &= 0.092\, \sigma^2_{\text{LS}} + 0.235\, \text{MSE}_{\text{p}} + P_{\text{mid}} + 0.0160\, C, \\
        E_{\text{far}}  &= 0.125\, \sigma^2_{\text{LS}} + 0.545\, \text{MSE}_{\text{p}} + P_{\text{far}} + 0.0240\, C,
    \end{align*}
    where all quantities depend on $(t, f)$.
    All values are finally clamped to a minimum numerical floor of $10^{-4}$.
\end{itemize}

\section{Channel Estimation with Known Covariance: Generated Algorithms}

\subsection{GPT-OSS~120B}
\label{sec:app_ce_cov_gptoss}
This algorithm implements an iterative Kronecker-structured LMMSE channel estimator. By pre-computing the eigen-decompositions of the time, frequency, and space covariance matrices, the algorithm efficiently projects a baseline noisy LS channel estimate into an uncorrelated eigen-domain. It then applies a two-pass adaptive Wiener filter iteratively. The first pass filters the signal in the eigen-domain using the global observation variance, while the second pass applies a localized correction in the original element domain. A damping factor ensures stable convergence across iterations, ultimately producing a highly refined channel estimate and its corresponding error variance.

Let $N_{\text{s}}$, $N_{\text{t}}$, and $N_{\text{f}}$ be the number of spatial antennas, time OFDM symbols, and frequency subcarriers, respectively. Let $H_{\text{LS}}(s, t, f)$ be the initial Least Squares channel estimate and $V_{\text{LS}}(s, t, f)$ be its corresponding error variance.

\subsubsection{Offline Pre-computation (Eigen-Decomposition)}
The algorithm assumes the full channel covariance matrix can be separated into three smaller Hermitian covariance matrices: $R_{\text{s}}$ (space), $R_{\text{t}}$ (time), and $R_{\text{f}}$ (frequency).
\begin{itemize}
    \item \textrm{Eigen-decomposition:} Each matrix is decomposed into its real eigenvalues and complex eigenvectors ($U^H$ denotes the conjugate transpose):
    \[ R_{\text{s}} = U_{\text{s}}\, \text{diag}(\Lambda_{\text{s}}) U_{\text{s}}^H \]
    \[ R_{\text{t}} = U_{\text{t}}\, \text{diag}(\Lambda_{\text{t}}) U_{\text{t}}^H \]
    \[ R_{\text{f}} = U_{\text{f}}\, \text{diag}(\Lambda_{\text{f}}) U_{\text{f}}^H \]
    \item \textrm{Prior Eigenvalue Product:} The combined 3D eigenvalue tensor is formed via the outer product:
    \[ \Lambda_{\text{prod}}(s, t, f) = \Lambda_{\text{s}}[s] \cdot \Lambda_{\text{t}}[t] \cdot \Lambda_{\text{f}}[f] \]
    \item \textrm{Marginal and Element Variances:} The diagonal of each covariance matrix represents the marginal variance, computed using the absolute squared eigenvectors:
    \begin{align*}
    \text{var}_{\text{s}} &= |U_{\text{s}}|^2 \Lambda_{\text{s}}, \\ 
    \text{var}_{\text{t}} &= |U_{\text{t}}|^2 \Lambda_{\text{t}}, \\
    \text{var}_{\text{f}} &= |U_{\text{f}}|^2 \Lambda_{\text{f}}
    \end{align*}
    where $|U|^2$ denotes the element-wise squared amplitude of $U$. The combined prior variance for each element in the original domain is:
    \[ \text{Var}_{\text{elem}}(s, t, f) = \text{var}_{\text{s}}[s] \cdot \text{var}_{\text{t}}[t] \cdot \text{var}_{\text{f}}[f] \]
\end{itemize}

\subsubsection{Initial Variance Transformation}
Before filtering, the element-domain LS error variance $V_{\text{LS}}$ is projected into the real-valued eigen-domain using tensor contractions to initialize $V_{\text{eig}}^{(0)}$:
\[ V_{\text{eig}}^{(0)} = |U_{\text{s}}|^2 |U_{\text{t}}|^2 |U_{\text{f}}|^2 V_{\text{LS}} \]

\subsubsection{Iterative Wiener Filtering}
Let $\epsilon$ be a small machine-precision constant added for numerical stability, $d \in (0, 1]$ be the damping factor, and $\tau_{\text{tol}}$ be the relative convergence tolerance. The channel state is initialized as $H^{(0)} = H_{\text{LS}}$. The algorithm loops over iterations $i = 0, 1, \dots$ through the following steps until the relative mean absolute change between $H^{(i)}$ and $H_{\text{ref}'}^{(i)}$ is less than $\tau_{\text{tol}}$ or the maximum number of iterations is reached:
\begin{itemize}
    \item \textrm{Transform to Eigen-domain:} 
    \[ H_{\text{eig}}^{(i)} = U_{\text{s}}^H U_{\text{t}}^H U_{\text{f}}^H H^{(i)} \]
    \item \textrm{First Wiener Pass (Eigen-domain):}
    \begin{align*} 
    W_1^{(i)} &= \frac{\Lambda_{\text{prod}}}{\Lambda_{\text{prod}} + V_{\text{eig}}^{(i)}}, \\
    H_{\text{eig}'}^{(i)} &= H_{\text{eig}}^{(i)} \cdot W_1^{(i)}, \\
    V_{\text{post}}^{(i)} &= \frac{\Lambda_{\text{prod}} \cdot V_{\text{eig}}^{(i)}}{\Lambda_{\text{prod}} + V_{\text{eig}}^{(i)}}.
    \end{align*}    
    \item \textrm{Transform to Element-domain:}
    \begin{align*}
    H_{\text{ref}}^{(i)} &= U_{\text{s}} U_{\text{t}} U_{\text{f}} H_{\text{eig}'}^{(i)}, \\
    V_{\text{post}'}^{(i)} &= |U_{\text{s}}|^2 |U_{\text{t}}|^2 |U_{\text{f}}|^2 V_{\text{post}}^{(i)}.
    \end{align*}
    \item \textrm{Second Wiener Pass (Element-domain):}
    \begin{align*} 
    W_2^{(i)} &= \frac{\text{Var}_{\text{elem}}}{\text{Var}_{\text{elem}} + V_{\text{post}'}^{(i)} + \epsilon}, \\
    H_{\text{ref}'}^{(i)} &= H_{\text{ref}}^{(i)} \cdot W_2^{(i)}.
    \end{align*}
    \item \textrm{State and Variance Update:}
    \[ V_{\text{final}}^{(i)} = V_{\text{post}}^{(i)} \cdot \left( \frac{\Lambda_{\text{prod}}}{\Lambda_{\text{prod}} + V_{\text{post}}^{(i)} + \epsilon} \right) \]
    Using the damping factor $d$, the states are updated for the next iteration:
    \begin{align*} 
    H^{(i+1)} &= d \cdot H_{\text{ref}'}^{(i)} + (1 - d) \cdot H^{(i)}, \\
    V_{\text{eig}}^{(i+1)} &= d \cdot V_{\text{final}}^{(i)} + (1 - d) \cdot V_{\text{eig}}^{(i)}.
    \end{align*}
\end{itemize}

\subsubsection{Final Output}
Let $I$ denote the final iteration index upon termination. The final eigen-domain variance $V_{\text{final}}^{(I)}$ is projected back into the element domain:
\[ V_{\text{final}'} = |U_{\text{s}}|^2 |U_{\text{t}}|^2 |U_{\text{f}}|^2 V_{\text{final}}^{(I)}. \]
The algorithm returns $H^{(I)}$ and $V_{\text{final}'}$.

\subsection{GPT~5.4}
\label{sec:app_ce_cov_gpt54}
This algorithm implements a separable, sequential LMMSE channel estimator. Instead of computing a computationally prohibitive joint filtering matrix across all dimensions, it utilizes independent, pre-computed covariance matrices for the frequency, time, and spatial domains. The algorithm dynamically generates regularized filtering matrices for each domain based on the current noise variance. These filters are then sequentially applied as linear transformations along their respective axes of the initial noisy LS channel estimate. To preserve edge details and avoid over-smoothing, the final channel estimate is constructed by taking a weighted linear combination of the initial LS estimate and the intermediate filtered states.

Let $N_0$ be the scalar noise variance representing the mean noise level (computed as the mean of the real part of the input noise tensor). Let $\hat{H}_{\text{LS}}$ be the initial LS channel estimate and $\sigma^2_{\text{LS}}$ be its corresponding error variance tensor. Let $R_{\text{f}}$, $R_{\text{t}}$, and $R_{\text{s}}$ be the pre-computed covariance matrices for the frequency, time, and space (receive antenna) dimensions, respectively.

\subsubsection{Filter Regularization Parameters}
The algorithm computes an axis-specific regularization parameter $\alpha$ (acting as an effective noise-to-signal ratio) that scales linearly with the input noise variance $N_0$:
\begin{align*}
    \alpha_{\text{f}} &= 0.0077 + 0.094 N_0 \\
    \alpha_{\text{t}} &= 0.0057 + 0.053 N_0 \\
    \alpha_{\text{s}} &= 0.0027 + 0.021 N_0
\end{align*}

\subsubsection{Covariance Normalization and Filter Construction}
For each covariance matrix $C \in \{R_{\text{f}}, R_{\text{t}}, R_{\text{s}}\}$ and its corresponding parameter $\alpha$, a specific filtering matrix $W$ is generated. 

First, the parameter $\alpha$ is clamped to a numerical floor: $\tilde{\alpha} = \max(\alpha, 10^{-6})$. 

Next, the covariance matrix is trace-normalized by its mean real diagonal element to ensure consistent scaling. Let $N$ be the dimension of the square matrix $C$:
\[ \tau = \max\left( \frac{1}{N} \sum_{i=1}^N C_{i,i}, 10^{-6} \right) \]
\[ \tilde{C} = \frac{1}{\tau} C \]
Then, the standard LMMSE filter matrix $W$ is constructed using the normalized covariance and the clamped regularization parameter, where $I$ is the identity matrix:
\[ W = \tilde{C} (\tilde{C} + \tilde{\alpha} I)^{-1} \]
This yields the three filtering matrices $W_{\text{f}}$, $W_{\text{t}}$, and $W_{\text{s}}$.

\subsubsection{Sequential Axis Filtering}
The filters are applied sequentially to the channel estimate tensor. Each filtering operation acts strictly along its targeted dimension. Let $\times_{\text{d}}$ denote the tensor mode-product (matrix multiplication along a specific axis).
\begin{itemize}
    \item \textrm{Frequency Filtering:}
    \[ H_{\text{f}} = W_{\text{f}} \times_{\text{f}} \hat{H}_{\text{LS}} \]
    \item \textrm{Time Filtering:}
    \[ H_{\text{t}} = W_{\text{t}} \times_{\text{t}} H_{\text{f}} \]
    \item \textrm{Spatial Filtering:}
    \[ H_{\text{s}} = W_{\text{s}} \times_{\text{s}} H_{\text{t}} \]
\end{itemize}

\subsubsection{Final State Blending and Variance Estimation}
The algorithm generates a composite final channel estimate $H_{\text{final}}$ by blending the original observation with the progressively smoothed intermediate states:
\[ H_{\text{final}} = 0.08 \hat{H}_{\text{LS}} + 0.12 H_{\text{f}} + 0.19 H_{\text{t}} + 0.61 H_{\text{s}} \]
Finally, the output error variance $E_{\text{final}}$ is modeled as a statically scaled version of the initial LS error variance, clamped to a numerical floor to maintain stability:
\[ E_{\text{final}} = \max(0.553 \cdot \sigma^2_{\text{LS}}, 10^{-6}) \]

\section{Link Adaptation: Generated Algorithms}

\subsection{GPT-OSS~120B}
\label{sec:app_la_gptoss}
This algorithm implements a Monte Carlo look-ahead link adaptation strategy based on a 1D Particle Filter. It tracks the posterior distribution of the unobservable SNR using a set of weighted particles that evolve via a random walk and are updated using binary ACK/NACK feedback. To select the optimal MCS, it evaluates the expected immediate and one-step future spectral efficiency for all candidate MCS values across the particle distribution, ensuring the average BLER remains below a specified target.

Let $N_{\text{particles}}$ be the total number of particles (set to 100). Let $s_i(t)$ denote the SNR hypothesis of particle $i$ at time step $t$, and $w_i(t)$ be its corresponding probability weight. Let $T_{\text{BLER}}$ be the target BLER. Let $\epsilon = 10^{-12}$ be a small constant for numerical stability. Let $p_{\text{cons}} = 0.2$ be the conservative percentile and $\Delta_{\text{margin}} = 0.5$ dB be the safety margin.

\subsubsection{Initialization}
If no history exists, the particles are initialized uniformly across the allowed SNR range $[\text{SNR}_{\text{min}}, \text{SNR}_{\text{max}}]$. All weights are initialized equally:
\[ w_i(0) = \frac{1}{N_{\text{particles}}} \]

\subsubsection{State Prediction (Random Walk)}
For each new observation step, the hidden SNR is assumed to evolve according to a Gaussian random walk:
\[ s_i(t) = s_i(t-1) + n_i \]
where $n_i$ is process noise drawn from a normal distribution $n_i \sim \mathcal{N}(0, \sigma^2)$ with standard deviation $\sigma = 0.5$. The updated SNR values are then clipped to remain within $[\text{SNR}_{\text{min}}, \text{SNR}_{\text{max}}]$.

\subsubsection{Measurement Update}
The weights are updated using Bayes' rule based on the binary feedback $f$ (where $f = 1$ for NACK, $f = 0$ for ACK) received for the previously chosen MCS $m$.
The likelihood $L_i$ of the observation given particle $i$'s SNR hypothesis is:
\begin{itemize}
    \item If NACK ($f = 1$): $L_i = \text{BLER}(s_i(t), m)$
    \item If ACK ($f = 0$): $L_i = 1 - \text{BLER}(s_i(t), m)$
\end{itemize}
To prevent numerical instability, the likelihoods are clamped: $L_i = \max(L_i, \epsilon)$.
The unnormalized weights are updated as:
\[ w_i^{\text{unnorm}} = w_i(t-1) \cdot L_i \]
The weights are then normalized so they sum to 1. If the sum of the unnormalized weights is zero or invalid, they are reset to a uniform distribution $w_i(t) = 1 / N_{\text{particles}}$. Otherwise:
\[ w_i(t) = \frac{w_i^{\text{unnorm}}}{\sum_{j} w_j^{\text{unnorm}}} \]

\subsubsection{Systematic Resampling}
To prevent particle degeneracy, the effective sample size $N_{\text{eff}}$ is calculated:
\[ N_{\text{eff}} = \frac{1}{\sum_i w_i(t)^2} \]
If $N_{\text{eff}} < N_{\text{particles}} / 2$, systematic resampling is triggered. Particles with high weights are duplicated, and particles with low weights are dropped. After resampling, all weights are reset to uniform: $w_i(t) = 1 / N_{\text{particles}}$.

\subsubsection{Candidate Feasibility Evaluation}
To select the next MCS, the algorithm computes the expected BLER for every possible candidate MCS $c$ across the current particle distribution:
\[ \mathbb{E}[\text{BLER}_c] = \sum_i w_i(t) \cdot \text{BLER}(s_i(t), c) \]
A candidate MCS $c$ is considered "feasible" only if its expected BLER is less than or equal to the target BLER $T_{\text{BLER}}$.

\subsubsection{One-Step Look-ahead Reward Maximization}
For each feasible candidate MCS $c$, the algorithm evaluates the expected total reward (throughput).

\begin{itemize}
    \item \textrm{Immediate Expected Reward:}
    \begin{equation*}
    R_{\text{immediate}}(c) = \sum_i w_i(t)  \,c \,
      \bigl(1 - \text{BLER}(s_i(t), c)\bigr)
    \end{equation*}
    
    \item \textrm{Future Expected Reward (Look-ahead):}
    The algorithm branches into two hypothetical futures (receiving an ACK or a NACK).
    \begin{align*}
    P_{\text{ACK}} &= \sum_i w_i(t) \bigl(1 - \text{BLER}(s_i(t), c)\bigr), \\
    P_{\text{NACK}} &= 1 - P_{\text{ACK}}
    \end{align*}
    It computes hypothetical future particle weights for both branches:
    \begin{align*} 
    w_i^{\text{ACK}} &= w_i(t) \cdot \frac{1 - \text{BLER}(s_i(t), c)}{P_{\text{ACK}}}, \\
    w_i^{\text{NACK}} &= w_i(t) \cdot \frac{\text{BLER}(s_i(t), c)}{P_{\text{NACK}}} 
    \end{align*}
    Using these distributions, the algorithm determines a ``safe'' future MCS for the ACK branch ($m_{\text{safe}}^{\text{ACK}}$) and the NACK branch ($m_{\text{safe}}^{\text{NACK}}$) using the conservative fallback strategy (see below). The expected future rewards are:
    \begin{align*}
    R_{\text{ACK}} &= \sum_i w_i^{\text{ACK}}  m_{\text{safe}}^{\text{ACK}}
      \bigl(1 - \text{BLER}(s_i(t), m_{\text{safe}}^{\text{ACK}})\bigr), \\
    R_{\text{NACK}} &= \sum_i w_i^{\text{NACK}}  m_{\text{safe}}^{\text{NACK}} \bigl(1 - \text{BLER}(s_i(t), m_{\text{safe}}^{\text{NACK}})\bigr)
    \end{align*}
    \item \textrm{Total Expected Reward:}
    \begin{align*}
    R_{\text{total}}(c) = R_{\text{immediate}}(c) + P_{\text{ACK}}  R_{\text{ACK}} + P_{\text{NACK}}  R_{\text{NACK}}
    \end{align*}
\end{itemize}
The algorithm selects and returns the candidate MCS $c$ that maximizes $R_{\text{total}}(c)$.

\subsubsection{Conservative Fallback}
If no candidate MCS satisfies the target BLER in Step 5 (or when determining safe future steps in Step 6), the algorithm defaults to a conservative estimate. It finds the $p_{\text{cons}}$-th percentile ($0.2$) of the current weighted SNR distribution, subtracts the $\Delta_{\text{margin}}$ ($0.5$ dB) safety margin, and returns the highest MCS that satisfies the target BLER $T_{\text{BLER}}$ for that specific pessimistic SNR value.

\subsection{GPT~5.4}
\label{sec:app_la_gpt54}
This algorithm implements a grid-based discrete Markov filter (histogram filter) paired with an asymmetric OLLA mechanism. It discretizes the SNR space into a fixed grid and tracks the probability distribution of the true SNR over time using a sliding window of up to 64 past transmissions. At each step, the distribution is propagated via Gaussian convolution (modeling a random walk) and updated through Bayesian inference based on ACK/NACK feedback. A trimmed mean of the posterior distribution yields a robust point estimate, discarding the extreme tails. Finally, the estimate is adjusted by an asymmetric OLLA offset---stepping up for ACKs and more aggressively down for NACKs---and a conservative warm-up margin is applied before selecting the MCS.

Let the target Block Error Rate be $T_{\text{BLER}}$. Let the history of observations be a sequence of $(f_i, m_i)$ pairs, where $f_i$ is the binary feedback ($1$ for NACK, $0$ for ACK) and $m_i$ is the transmitted MCS at step $i$. The algorithm processes an observation window of length $N_{\text{hist}} \le 64$. If the history of observations is empty ($N_{\text{hist}} = 0$), the algorithm bypasses all tracking and directly returns the highest MCS index satisfying the target BLER for a hardcoded estimate of $7.0$ dB.

\subsubsection{Grid Setup and Prior Initialization}
The continuous SNR space is discretized into a fixed grid $G$ of $N_\text{grid} = 169$ points, linearly spaced from $-12.0$ dB to $30.0$ dB. The spacing between points is $dx = 0.25$ dB.
A pre-computed lookup table, $\text{BLER}_{\text{table}}(s, m)$, stores the BLER for every grid point $s \in G$ and MCS $m$.

The initial probability distribution $P_0(s)$ over the grid is set to a normalized discrete Gaussian prior centered at $\mu_{\text{prior}} = 9.0$ dB with a standard deviation of $\sigma_{\text{prior}} = 6.5$ dB. The active tracking distribution is initialized as $P(s) \leftarrow P_0(s)$, defined as:
\[ P_0(s) = \frac{\exp\!\left( -0.5 \left(\frac{s - \mu_{\text{prior}}}{\sigma_{\text{prior}}}\right)^2 \right)}{\sum_{x \in G} \exp\!\left( -0.5 \left(\frac{x - \mu_{\text{prior}}}{\sigma_{\text{prior}}}\right)^2 \right)} \]

\subsubsection{Bayesian Grid Tracking}
For each observation $i \in \{1, \dots, N_{\text{hist}}\}$ in the history window, the tracking distribution $P(s)$ is sequentially updated in-place through four steps:

\begin{itemize}
\item \textrm{Prediction (Convolution):}
The state undergoes a random walk modeled by convolving the distribution with a discrete, normalized Gaussian kernel $K$ with standard deviation $\sigma_{\text{rw}} = 0.52$ dB:
\begin{align*} 
K(\text{offset}) &= \exp\!\left( -0.5 \left(\frac{\text{offset} \cdot dx}{\sigma_{\text{rw}}}\right)^2 \right) \\
P(s) &\leftarrow (P * K)(s)
\end{align*}

\item \textrm{Forget Mix (Regularization):}
To prevent assigning absolute zero probability to any state, a fraction $\alpha = 0.006$ of a uniform distribution is mixed in:
\[ P(s) \leftarrow (1 - \alpha) P(s) + \frac{\alpha}{N_\text{grid}} \]

\item \textrm{Measurement Update:}
The probabilities are scaled by the likelihood $L_i(s)$ of the observation $f_i$ given the transmitted MCS~$m_i$. The likelihood is clamped to $[10^{-7}, 1 - 10^{-7}]$:
\begin{align*}
L_i(s) &= \begin{cases} 
      \text{BLER}_{\text{table}}(s, m_i) & \text{if NACK } (f_i = 1) \\
      1 - \text{BLER}_{\text{table}}(s, m_i) & \text{if ACK } (f_i = 0) 
   \end{cases} \\
P(s) &\leftarrow P(s) \cdot L_i(s)
\end{align*}

\item \textrm{Normalization:}
The distribution is normalized. If the sum $S = \sum_{s \in G} P(s)$ is zero or invalid, the filter resets:
\[ P(s) \leftarrow \begin{cases} 
      P(s) / S & \text{if } S > 0 \\
      P_0(s) & \text{otherwise}
   \end{cases} \]
\end{itemize}

\subsubsection{Posterior Trimmed Mean}
After processing the history, the algorithm extracts a single representative SNR value from the final updated distribution $P(s)$. Instead of a standard mean, it computes a trimmed mean to ignore misleading probability tails:
\begin{itemize}
    \item Compute the cumulative distribution function, $\text{CDF}(s_k)$, of $P$ over the grid points $s_k \in G$.
    \item Let $k_{\text{lo}}$ and $k_{\text{hi}}$ be the integer indices of the grid corresponding to the 15th percentile ($q_{\text{lo}} = 0.15$) and 85th percentile ($q_{\text{hi}} = 0.85$) of the CDF, respectively.
    \item The base SNR estimate is the weighted center of mass strictly between these two bounds:
    \[ S_{\text{base}} = \frac{\sum_{k=k_{\text{lo}}}^{k_{\text{hi}}} P(s_k) \cdot s_k}{\sum_{k=k_{\text{lo}}}^{k_{\text{hi}}} P(s_k)} \]
    \item If the trimmed weight sum is invalid or zero, it falls back to the median (50th percentile) of $P$.
\end{itemize}

\subsubsection{Asymmetric OLLA Offset}
An OLLA offset is accumulated over the entire available history, stepping the SNR estimate up slightly for every ACK and down more aggressively for every NACK.
\begin{itemize}
    \item Up-step (for ACKs): $\text{step}_\text{up} = 0.0237$ dB
    \item Down-step (for NACKs): $\text{step}_\text{down} = \text{step}_\text{up} \cdot \frac{1 - T_{\text{BLER}}}{\max(T_{\text{BLER}}, 10^{-9})}$
\end{itemize}
With total ACKs $N_{\text{ACK}}$ and total NACKs $N_{\text{NACK}}$, the raw offset is calculated as:
\begin{equation*}
\text{Offset}_{\text{raw}} = \text{step}_\text{up} \cdot N_{\text{ACK}} - \text{step}_\text{down} \cdot N_{\text{NACK}}
\end{equation*}
This offset is then clamped to $[-2.94,\; 0.25]$~dB, yielding $\text{Offset}_\text{clamped}$.

\subsubsection{Warm-up Margin and Final Selection}
When the observation history is short and the filter is not yet converged, a conservative warm-up margin is applied based on the total number of observations $n$:
\begin{itemize}
    \item $n < 5$ observations: $\text{Margin} = -0.95$ dB
    \item $n < 10$ observations: $\text{Margin} = -0.35$ dB
    \item Otherwise: $\text{Margin} = 0.0$ dB
\end{itemize}
The final estimated SNR is the sum of the base trimmed mean, the OLLA offset, and the warm-up margin:
\[ S_\text{final} = S_\text{base} + \text{Offset}_\text{clamped} + \text{Margin} \]
$S_\text{final}$ is clamped to the grid boundaries $[-12.0,\; 30.0]$~dB.

The algorithm then uses the pre-computed BLER curves to return the highest MCS index whose expected BLER at $S_\text{final}$ is $\le T_{\text{BLER}}$.

\end{document}